\pdfoutput=1
\documentclass[11pt]{article}

\usepackage[]{acl}

\usepackage{times}
\usepackage{latexsym}

\usepackage[T1]{fontenc}
\usepackage[utf8]{inputenc}

\usepackage{microtype}

\usepackage{booktabs}
\usepackage{multirow}
\usepackage{rotating}
\usepackage{paralist}
\usepackage{comment} 
\usepackage[normalem]{ulem}

\usepackage{tipa}

\title{A Question-Answer Driven Approach to Reveal Affirmative Interpretations\\from Verbal Negations}

\author{Md Mosharaf Hossain\mbox{\normalfont,}\textsuperscript{\textipa{8}}
        Luke Holman\mbox{\normalfont,}\textsuperscript{\textipa{U}}  
        Anusha Kakileti\mbox{\normalfont,}\textsuperscript{\textipa{8}}  
        Tiffany Iris Kao\mbox{\normalfont,}\textsuperscript{\textipa{7}} \\
        \textbf{Nathan Raul Brito}\mbox{\normalfont,}\textsuperscript{\textipa{8}} 
        \textbf{Aaron Abraham Mathews}\mbox{\normalfont,}\textsuperscript{\textipa{8}} \and       
        \textbf{Eduardo Blanco}\textsuperscript{\textipa{U}}\\
\textsuperscript{\textipa{8}}University of North Texas \hspace{.2cm}
\textsuperscript{\textipa{U}}Arizona State University \hspace{.2cm}
\textsuperscript{\textipa{7}}University of Texas at Austin\\
{\footnotesize
\texttt{\{mdmosharafhossain,nathanbrito,aaronmathews\}@my.unt.edu} \hspace{.2cm}
\texttt{lholman2@asu.edu} \hspace{.2cm}
}
\\
{\footnotesize
\texttt{akakileti@gmail.com} \hspace{.2cm}
\texttt{tiffanyiris1004@gmail.com} \hspace{.2cm}
\texttt{eduardo.blanco@asu.edu}
}
}

\begin{document}
\maketitle
\begin{abstract}
This paper explores a question-answer driven approach to reveal affirmative interpretations from verbal negations (i.e., when a negation cue grammatically modifies a verb).
We create a new corpus consisting of 
4,472 verbal negations 
and discover that 67.1\% of them 
convey that an event actually occurred. 
Annotators generate and answer 7,277 questions % converted for 4,000
for the 3,001 negations 
that convey an affirmative interpretation.
We first cast the problem of revealing affirmative interpretations from negations as a natural language inference (NLI) classification task.
Experimental results show that state-of-the-art transformers trained with existing NLI corpora are insufficient to reveal affirmative interpretations.
We also observe, however, that fine-tuning brings small improvements.
In addition to NLI classification, we also explore the more realistic task of generating affirmative interpretations directly from negations with the T5 transformer.
We conclude that the generation task remains a challenge as T5 substantially underperforms humans.
\end{abstract}

\section{Introduction}
\label{s:introduction}
%The negation phenomenon is common to all human languages. 
Negation can be understood as an operator that transforms the meaning of some expression into another expression whose meaning is in some way opposed to the original expression~\cite{sep-negation}.
%Negation of an expression occurs when it is linked to another expression that is in some way opposed to that of the original expression \cite{horn2015negation}. 
Typically, negated statements are
less informative than affirmative statements (e.g., ``Paris is not located in England'' vs. ``Paris is located in France'').
Negated statements are also harder to process and understand by humans~\cite{sep-negation}.
%It is also harder to process from a cognitive and harder to process its semantics \cite{horn2015negation}.
According to \newcite{horn1989natural},
negations carry affirmative meanings.
These underlying affirmative meanings, which we refer to as \emph{affirmative interpretations}, 
range from implicatures to entailments. 
%According to \citeauthor{horn1989natural}, negations expose affirmative meanings, which range from implicature to entailment. 
For example, the negated statement
(1) ``Mary never drives long distances without a full tank of gas",
carries at least the following affirmative interpretations: 
(1a) ``Mary drives long distances,''
(1b) ``Mary fills the gas tank before starting a long drive,'' and 
(1c) ``Mary might drive short and medium distances without a full tank of gas.''
%As we shall see,
%revealing this kind of intuitive positive interpretations from negation is challenging for state-of-the-art models.

% Start: Table 
\begin{table}
\small
\centering
\begin{tabular}{p{1.69in}p{1.0in}}
\toprule
%Example  \\ 
%\midrule
\multicolumn{2}{p{2.69in}}{An extinct volcano is one that has \underline{not} \emph{erupted} in recent history.} \\ \midrule
    - Did something erupt?      & \emph{Yes} \\
    - What erupted?             & \emph{An extinct volcano} \\
    - When did something erupt? & \emph{In the past} \\
    \multicolumn{2}{p{2.69in}}{Affirm. Intp: \emph{An extinct volcano erupted in the past.}} \\
%\midrule
%    May something erupt? - Yes \\
%    What may erupt? - An active volcano \\
%    When may something erupt? - in recent history \\
%    Affirm. Intp.: An active volcano may erupt in recent history. \\
\bottomrule 
\addlinespace
          
\toprule
\multicolumn{2}{p{2.69in}}{It was \underline{not} \emph{formed} by a natural process.}  \\ \midrule
   - Was something formed?         & \emph{Yes} \\
   - What was formed?              & \emph{It}  \\
   - What was something formed by? & \emph{An artificial process} \\
   \multicolumn{2}{p{2.69in}}{Affirm. Intp: \emph{It was formed by an artificial process.}} \\ 
%\midrule

%(3) Most of these sound waves did \underline{not} \emph{hit} submarines.   \\ 
%a. \emph{A few of these sound waves hit submarines.} \\ 
%(What hit something? - A few of these sound waves; What did something hit? - submarines) \\

\bottomrule
\end{tabular}
\caption{Sentences containing negation,
  questions and answers about the affirmative counterpart of the main event,
  and
  the underlying affirmative interpretation.
%  We investigate methods to reveal affirmative interpretations from negations.
  %Example sentences containing negation and their affirmative interpretations (Affirm. Intp.). 
%The \emph{yes/no} question checks whether an event is actually occurs (or occurred). 
%The wh-questions reveal the arguments of the event that eventually produce an affirmative interpretation.
}
\label{t:motivational-examples}
\end{table}
% End: Table 

In order to empower models to comprehend negation,
most previous works target scope~\cite{vincze2008bioscope,L12-1077} and focus~\cite{blanco-moldovan:2011:ACL-HLT20111} detection (Section \ref{s:related-work}).
Scope refers to the part of the meaning that is negated and focus refers to the part of the scope that is most prominently or explicitly negated \cite{english.grammar.2002}. 
Scope and focus detection plays a crucial role to understand what part of a negated statement is actually negated.
These tasks do not, however, reveal affirmative interpretations---they tag tokens as belonging or not belonging to the scope and focus of a negation.
%These, it remains a challenge to evince an intended meaning. 

In this paper, we present a question-answer driven approach to reveal affirmative interpretations from verbal negations (i.e., when a negation cue grammatically modifies a verb).
We adapt QA-SRL \cite{he-etal-2015-question,fitzgerald-etal-2018-large}
to collect questions and answers regarding the arguments of the affirmative counterpart of a negated predicate.
Then, we manipulate the questions and answers to generate an affirmative interpretation.
We find that generating and answering questions is intuitive to non-experts (albeit they are native English speakers).
Consider the examples in Table \ref{t:motivational-examples}.
Annotators first generate and answer a question regarding whether the main predicate in the sentence occurred (with unknown arguments at this point).
Then, they generate and answer questions about the arguments of the affirmative counterpart of the main predicate.
Arguments may come directly from the negated statement (e.g., What erupted? \emph{An extinct volcano})
or using commonsense and world knowledge after reading the negated statement
(e.g., When did something erupt? \emph{In the past}).
After collecting questions and answers, we automatically generate an affirmative interpretation in the form of a statement (e.g., \emph{An extinct volcano erupted in the past}).

The main contributions of this paper are:\footnote{
Corpus and code available at \url{https://github.com/mosharafhossain/AFIN}.}%\tbd{I am not sure about the footnote with the English translation. MH: I think thisis alright }
\begin{compactenum}
\item A question-answer driven annotation schema to create AFIN,
%\footnote{\emph{Afín} is an adjective in Spanish. English translation: nerby, adjacent; related, similar.}
a corpus of verbal negations and their AFfirmative INterpretations (4,472 negations, 7,277 questions and answers, and 3,001 affirmative interpretations); 
\item Corpus analysis indicating which predicate arguments are most often rephrased in the affirmative counterparts;
\item Casting the problem of revealing affirmative interpretations as a natural language inference task and showing that it is challenging for state-of-the-art transformers; and
\item Casting the problem of revealing affirmative interpretations as a generation task and showing that the T5 transformer substantially underperforms humans.
\end{compactenum}
%task, where the question-answer pairs represent predicate-argument structure. 
%We observe that answering questions is easy for non-experts, and this approach intuitively identifies the events having affirmative interpretations as well as finds the arguments to generate them.
%Consider the negated sentence (1) in Table \ref{t:motivational-examples}, 
%where annotators initially ask a \emph{yes/no} question ``Did something erupted?", to check whether the event (i.e., \emph{erupted}) actually occurred. If the answer to above question is \emph{Yes}, then they ask a few additional questions to reveal the arguments (e.g., ``What erupted?" - ``An extinct volcano"; ``When did something erupt?" - ``in the past"). 
%The arguments as a whole refer to the complete affirmative interpretation ``An extinct volcano erupted in the past", which we produce by applying some simple rules (Section \ref{ss:qa-to-pi}). 
%Additionally, the same negated event (i.e., erupted) can reveal an alternative affirmative interpretation such as ``An active volcano may erupt in recent history." 
%In case an event expresses no occurrence or action, the first question is answered with \emph{No}, indicating that there is no affirmative interpretation for the event (e.g., ``They do \underline{not} \emph{contain} oxygen"; ``Do they contain something?" - ``No").  

\section{Related Work}
\label{s:related-work}

Revealing affirmative interpretations from negations is a challenging endeavor.
In the literature, researchers primarily seek to identify scope and focus of negation. 
The creation of the BioScope \cite{Szarvas:2008:BCA:1572306.1572314} and ConanDoyle-Neg~ \cite{L12-1077} corpora spearheaded research on scope detection~\cite{morante-daelemans-2009-metalearning}.
Proposals include using
traditional machine learning~\cite{lapponi-etal-2012-uio},
off-the-shelf semantic parsers and semantic representations~\cite{packard-etal-2014-simple}, and
neural networks~\cite{fancellu-etal-2016-neural,fancellu-etal-2017-detecting}.
PB-FOC~\cite{blanco-moldovan:2011:ACL-HLT20111} is the largest corpus with focus of negation annotations.
Recent proposals for focus detection include graph-based models with discourse information~\cite{zou-etal-2014-negation, zou-etal-2015-unsupervised},
neural networks with word-level and topic-level attention~\cite{shen-etal-2019-negative},
and networks using scope information and context~\cite{hossain-etal-2020-predicting}.
Scope and focus are useful to identify what is and what is not negated in a negated statement.
Consider the second example in Table~\ref{t:motivational-examples}.
Scope and focus do reveal that \emph{It was formed}---everything but the focus (i.e., \emph{by a natural process}) is affirmative---but provide no hints about \emph{how} it was formed (i.e., \emph{by an artificial process, artificially, etc)}.
The main goal of this paper is to find these affirmative counterparts to generate affirmative interpretations.
%One could describe our affirmative interpretations as inferences understood when reading negated statements.

More related to the work presented here,
\newcite{sarabi-etal-2019-corpus} present a corpus of negations and their underlying affirmative interpretations (they call them positive interpretations).
We are inspired by them but bypass several of their limitations. 
First, they only work with negations from Simple Wikipedia,
a site devoted to English learners.
As a result, their corpus uses (relatively) unsophisticated vocabulary and grammar.
Second, they impose several restrictions on the negations they work with
(e.g., negation cue modifies root verb,
sentences between 6 and 25 tokens and not including certain tokens (because, until, etc.)).
Third, their affirmative interpretations are restricted to a rephrasing of the statement containing negation with only one change: an argument of the negated predicate.
In contrast, we barely impose restrictions on the negations we work with (no questions and no auxiliary verbs). 
More importantly, we introduce a question-driven approach that allow us to obtain multiple affirmative interpretations with increasing degrees of complexity (see examples in Table \ref{t:qas-to-ai}).

Recently, \newcite{jiang-etal-2021-im} study the problem of identifying commonsense implications of negations and contradictions.
More specifically, 
they work with if-then rules such as 
\emph{If X does not learn new things, then X does not gain new knowledge} and
\emph{If X does not leave the building, then X stays in the building}.
These rules capture general commonsense knowledge about what happens if an event does not occur.
Unlike them, we work with naturally occurring sentences that include negated predicate-argument structures with many arguments (agent, theme, manner, time, etc.). 
%\tbd{\textcolor{red}{MH: deleted the argument ``location" to save space.}}
In addition, our affirmative interpretations reveal that predicates that are grammatically negated are actually factual (but with different arguments).

% Start: Table 
\begin{table*}[t!]
\small
\centering
\begin{tabular}{p{1.0in} c cccccccc}
\toprule
       &   Sent.    & WH         &   AUX  &   SUB        &  VERB     &   OBJ1     &   PREP  &   OBJ2 & \\ \midrule
\multirow{3}{*}{Predicate questions}
       &  (a)  &            &   Was  &   something  &  formed  &            &         &               & ? \\ 
       &  (b)  &            &   Does &   something  &  kill    & someone    &         &               & ? \\ 
       &  (c)  &            &   Will &   someone    &  have    &            &    to   &  do something & ? \\ \midrule

\multirow{3}{*}{Argument questions}
       &  (a)  & What       &   was  &   something  &  formed  &            &   by    &              & ? \\ 
       &  (b)  & How often  &   does &   something  &  kill    &   someone  &         &              & ? \\ 
       &  (c)  & Who        &   will &              &  have    &            &   to    & do something & ? \\

\bottomrule
\end{tabular}

\begin{comment}
\begin{tabular}{ll cccccccc}
\toprule
&& WH         &   AUX  &   SUB        &  VERB     &   OBJ1     &   PREP  &   OBJ2 & \\ \midrule
\multirow{4}{*}{Predicate questions}
%&&            &   Did  &   something  &  erupt   &            &         &               & ? \\ 
&&            &   Was  &   something  &  formed  &            &         &               & ? \\ 
&&            &   Does &   something  &  kill    & someone    &         &               & ? \\ 
&&            &   Will &   someone    &  have    &            &    to   &  do something & ? \\ \midrule

\multirow{4}{*}{Argument questions}
%&& What       &        &              &  erupted &            &         &              & ? \\ 
&& What       &   was  &   something  &  formed  &            &   by    &              & ? \\ 
&& How often  &   does &   something  &  kill    &   someone  &         &              & ? \\ 
&& Who        &   will &              &  have    &            &   to    & do something & ? \\

\bottomrule
\end{tabular}
\end{comment}
\caption{Predicate questions and argument questions (one per negated predicate) generated by annotators from the sentences 
%(a) \emph{Scientists think that it will probably \underline{not} \emph{erupt} again.},
(a) \emph{It was \underline{not} \emph{formed} by a natural process.},
(b) \emph{However, the ground shaking almost \underline{never} \emph{kills} people, [\ldots]}, and
%and the ground does not swallow someone up.}, and
(c) \emph{[\ldots], he hopes Australian teams will \underline{not} \emph{have} to travel so much to meet first class competition}.
% By building up the teams in the Asia Oceania Zone, he hopes Australian teams will \underline{not} %\emph{have} to travel so much or so far to meet first class competition
}
\label{t:using-slots}
\end{table*}
% End: Table 

\section{A Question-Answer Driven Approach to Collect Affirmative Interpretations}
\label{s:qa-driven-data-annotation}
This section outlines our approach to create AFIN, a corpus of verbal negations and their affirmative interpretations. We first describe the sources of naturally occurring negations in our corpus.
Then, we outline the template-based approach to guide annotators in generating and answering questions about the \emph{affirmative counterpart} of the negated predicate.
Lastly, we describe the process to generate natural-language affirmative interpretations from the questions and answers.

\subsection{Collecting Sentences Containing Negation}
\label{ss:selecting-source-corpus}

\begin{figure}[t!]
\centering
\includegraphics[width=0.97\columnwidth]{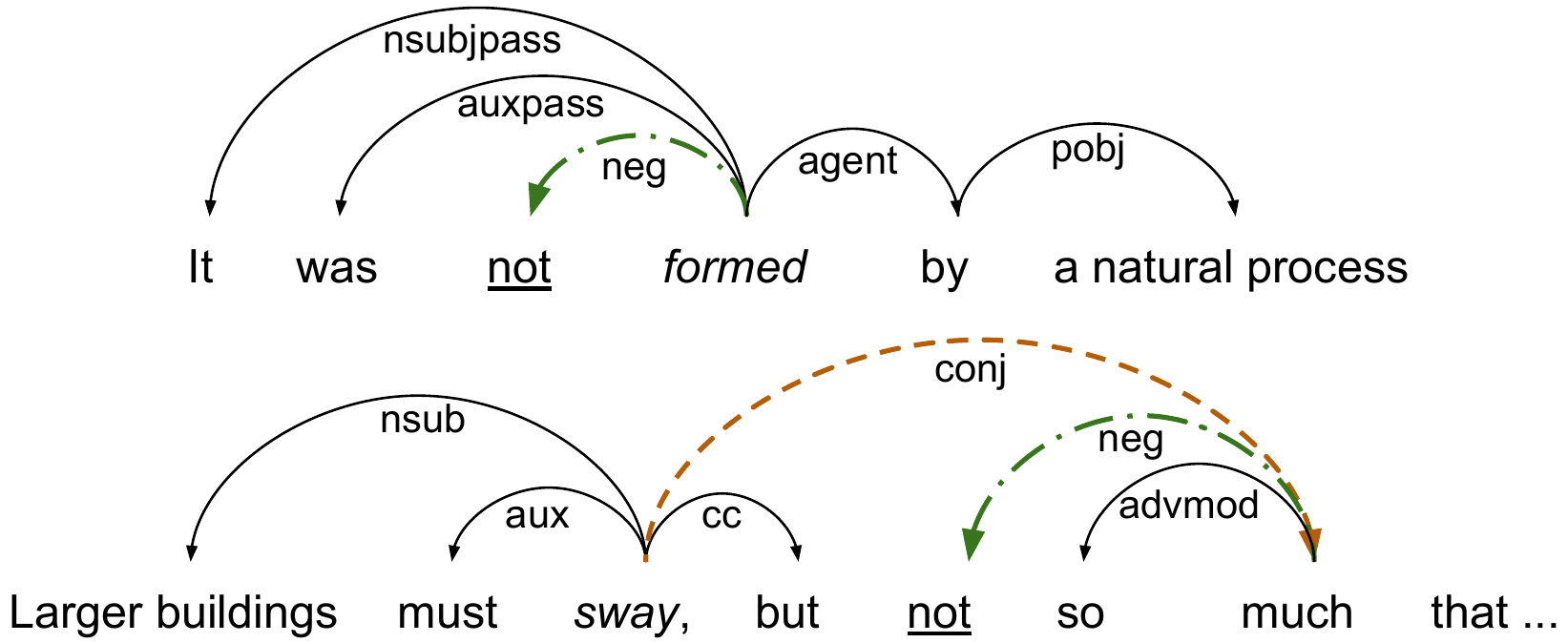} % Reduce the figure size so that it is slightly narrower than the column. Don't use precise values for figure width.This setup will avoid overfull boxes.
\caption{Illustration of the criteria to select negated verbs.
  We select all negations that modify non-auxiliary verbs either
  directly (top) or
  indirectly (bottom).}
  %Negated event identification in two scenarios: 
%(a) negation cue directly modifies a verb (\emph{not} modifies \emph{formed}) and 
%(b) negation cue indirectly modifies a verb (\emph{not} modifies \emph{much}, then \emph{much} modifies \emph{sway}).}
\label{f:negated-verb-identify}
\end{figure}

We start with the sentences in QA-SRL Bank 2.0~\cite{fitzgerald-etal-2018-large},
a corpus with 64,000 
sentences across three domains: Wikipedia, Wikinews, and science textbooks~\cite{kembhavi2017you}. 
Motivated by 
\newcite{fancellu-etal-2016-neural}, 
we select sentences containing negations checking for the following negation cues:
\emph{not}, \emph{n't}, \emph{no}, \emph{never}, \emph{without}, \emph{nothing}, \emph{none}, \emph{nobody}, \emph{nowhere}, and \emph{neither} and \emph{nor}.
%\tbd{\textcolor{red}{MH: Removed the term "copula" from here. Copula has many types. We excluded the auxiliary type of copula.}}
We only impose two restrictions:
the sentences cannot be questions and the negation cues have to modify a verb that is not an auxiliary verb. 
We check the latter using universal dependencies as extracted by the parser in spaCy~\cite{spacy}.
We consider cues that directly or indirectly modify the verb, as exemplified in~Figure~\ref{f:negated-verb-identify}.
%\tbd{\textcolor{red}{MH: removed a line to save some space. Please feel free to change.}}
%Our motivation to disregard negated copula verbs is that the resulting affirmative interpretations are (arguably) less interesting (e.g., \emph{Mary is calm} from \emph{Mary is not stressed out}.
We will use \emph{target} verb to refer to the negated verb in the remaining of the paper.

%, where the cue \emph{not} directly modifies \emph{formed}
%, or indirectly modifies a verb (e.g., in Figure~\ref{f:negated-verb-identify}(b)).
%, \emph{not} modifies \emph{much}, then \emph{much} modifies the verb \emph{sway}
%Furthermore, we exclude the verbs that are marked as auxiliary (e.g., is, was, be) by the dependency parser. 

%\tbd{predicates or verbs}

\subsection{Generating and Answering Questions}
\label{ss:annotation-task-design}
Given a sentence and a target verb,
our goal is to guide annotators to generate and answer questions about the (potential) affirmative counterpart of the target verb.
First, they ask a \emph{predicate question} to determine whether the affirmative counterpart of the target verb is factual (with unknown arguments).
If it is, then they ask and answer \emph{argument questions} about the arguments of the affirmative counterpart of the target verb.
Consider the following sentence:
\emph{However, \underline{no} children resulted from the marriage}.
The answer to the predicate question (Did anything result?) is \emph{No}, thus no argument questions are considered.
Now consider another sentence: \emph{Cloning does not happen naturally}.
The answer to the predicate question (Does something happen?) is \emph{Yes}, thus annotators continue asking and answering argument questions: What happens? \emph{Cloning} and How does it happen? \emph{Artificially} (or \emph{with human intervention}, for example).

\noindent
\textbf{Template-Based Question Generation}
\label{sss:templates}
In principle we could allow annotators to generate questions following their preferred wording.
We found, however, that guiding them increases consistency and speed.
To this end, we adapt the seven-slot template technique by \newcite{he-etal-2015-question}.
For predicate questions (expected answer: Yes or No), we use the following combinations of slots:
AUX x SUB x VERB x OBJ1 x PREP x OBJ2.
For argument questions, we include an additional slot in the first position: WH.
The full list of values for each slot are detailed in Appendix \ref{s:additional-question-template}. 
We provide below some examples for each slot.  

\begin{compactitem}
\item WH: Who, What, Whom, When, Where, etc.
\item AUX: is, was, does, did, has, had, can, etc.
\item SUB: something, or someone
\item VERB: full conjugation of the target verb
\item OBJ1: something, or someone
\item PREP: by, to, for, with, about, of, or from
\item OBJ2: someone, something, somewhere, do, doing, etc.
\end{compactitem}

The templates allow annotators to generate a wide variety of questions.
Table \ref{t:using-slots} shows several examples of predicate and argument questions generated from three target verbs.
Note that humans are needed to choose values for each slot so that the resulting question is correct
(right auxiliaries, conjugation, tense, number matching, etc.).
Annotators generate questions in the following order of wh-words:
\textit{who} (or \textit{what}) does/did  
(something) to 
\textit{whom} (or \textit{what}), 
\textit{when}, 
\textit{where}, 
\textit{how}, \textit{how much}, \textit{how many}, \textit{how long}, \textit{how often}, and
%\footnote{Along with \textbf{how}, we include a few other common options such as \textbf{how much}, how many}, \textbf{how long}, and \textbf{how often}.}, and 
\textit{why}.
This order makes the generation of affirmative interpretations in natural language easier (Section \ref{ss:qa-to-pi}).
%, to aid in generating an affirmative interpretation automatically from the questions and answers (see Section \ref{ss:qa-to-pi}).

\begin{table*}[th!]
\small
\centering
\begin{tabular}{p{0.09in} cccccc l l}
\toprule
& WH         &   AUX   &   SUB        &  VERB     &      PREP  &&  Answer & Affirmative Interpretation \\ \midrule
\multirow{2}{*}{(1)} & What       &         &              &  flows   &            &  ?   & Lava               & Lava flows.  \\ 
& Where      &   does  &  something   &  flow    &            &  ?   & Close to vents     & Lava flows close to vents. \\  
\midrule

\multirow{3}{*}{(2)} & What       &  was    &              &  classified   &        &  ?   & Fungi             & Fungi were classified. \\ 
& What       &  was    &  something   &  classified   &   as   &  ?   & Plants            & Fungi were classified as plants. \\ 
& When       &  was    &  something   &  classified   &        &  ?   & In the past       & Fungi were classified as plants in the past. \\

\bottomrule
\end{tabular}

\begin{comment}
\begin{tabular}{l cccccc lll ll}
\toprule
%\multicolumn{10}{l}{\emph{The steep sides form because the lava can \underline{not} \emph{flow} too far from the vent}.} \\ \addlinespace
%& \multicolumn{6}{c}{Question} && \multirow{2}{*}{Answer} && \multirow{2}{*}{Affirmative Interpretation} \\ \cmidrule{2-6}
& WH         &   AUX   &   SUB        &  VERB     &      PREP  &&&&  Answer && Affirmative Interpretation \\ \midrule
%\multirow{3}{*}{\parbox{3.5cm}{(1) The steep sides form because the lava can \underline{not} \emph{flow} too far from the vent.}}
\multirow{2}{*}{(1)} & What       &         &              &  flows   &            &  ?   &&& Lava               && Lava flows.  \\ 
& Where      &   does  &  something   &  flow    &            &  ?   &&& Close to vents     && Lava flows close to vents. \\ 
%&            &         &              &          &            &      &                    &  \\ 
\midrule

%\multirow{3}{*}{\parbox{3.5cm}{ (2) Today, fungi are \underline{no} longer \emph{classified} as plants.}}
\multirow{3}{*}{(2)} & What       &  was    &              &  classified   &        &  ?   &&& Fungi             && Fungi were classified. \\ 
& What       &  was    &  something   &  classified   &   as   &  ?   &&& Plants            && Fungi were classified as plants. \\ 
& When       &  was    &  something   &  classified   &        &  ?   &&& In the past       && Fungi were classified as plants in the past. \\

\bottomrule
\end{tabular}
\end{comment}
\caption{Examples of questions and answers generated by annotators and the resulting affirmative interpretations.
The sentences containing the negated predicates are
(1) \emph{The steep sides form because the lava can\underline{not} \emph{flow} too far from the vent.}
and
(2) \emph{Today, fungi are \underline{no} longer \emph{classified} as plants.}
We do not show the OBJ1 and OBJ2 slots because they are empty for all the questions in these examples.
%Sample examples showing how we go from question-answer pairs to a complete affirmative interpretation. 
%In the last column, the last sentence under each event refers to the complete affirmative interpretation. 
%While generating affirmative interpretation from QAs, the slots OBJ1 and OBJ2 are not required (and are not shown here).
}
\label{t:qas-to-ai}
\end{table*}

\noindent
\textbf{Answering Questions and Assigning Confidence Scores}
\label{sss:correctness-scoring}
Immediately after generating a question (i.e., before generating the next question),
annotators answer it and indicate how confident they are in their answer.
Note that several compatible answers are usually possible
(e.g., \emph{before} and \emph{in the past} are usually interchangeable).
Answers may come from the sentence containing the target verb and its arguments,
or written by annotators using commonsense and world knowledge.
Consider the following sentence: \emph{The steep sides form because the lava cannot flow too far from the vent} (example (1) in Table \ref{t:qas-to-ai}).
The answer to \emph{What flows?} comes from the sentence: \emph{Lava}.
On the other hand, the answer to \emph{Where does something flow?} is a rewrite of an argument of the target verb: \emph{close to vents}. 
%\tbd{\textcolor{red}{MH: replaced ``containing" with ``with" to save a line.}}
In the second example of Table \ref{t:qas-to-ai}, all answers come from the sentence with the target verb except \emph{When was something classified?}, which is \emph{In the past}.

%Given the context in the sentence, more than one answer is possible to an argument question. It is likely that some answers are certainly true, but others might be more difficult to determine. 
%We ask annotators to answer with their best judgment to the questions they write. 
%As well, we allow the annotators to grade their answers according to a score scale derived from \citeauthor{sarabi-etal-2019-corpus}.
%Definitions with examples are below (more examples in supplementary material).
Regardless of where answers come from, annotators assign a confidence score
 %\tbd{quotations vs. italics not consistent?. MH: seems alright}
using a four-point Likert scale:
\begin{compactitem}
    \item 4: Extremely confident.
      I am certain that the answer is correct given the negated statement.
      For example, given ``Scientists think that it will probably \underline{not} \emph{erupt} again,''
      annotations answer \emph{When did something erupt?} with \emph{In the past} and assign a score of 4.
 
    \item 3: Very confident.
      My answer is very likely correct given the negated statement.
      For example, given ``These volcanoes usually do \underline{not} \emph{produce} streams of lava,''
      an annotator generated \emph{How often does something produce?} and answered \emph{Rarely} with a confidence score of 3.
      %      the answer to The annotated answer has a good chance to be true. For example, in ``These volcanoes usually do \underline{not} \emph{produce} streams of lava", the answer ``rarely" is a probable response to the question ``How often does something produce?".

    \item 2: Moderately confident.
      My answer is likely correct given the negated statement.
      There are, however, many possible answers and my answer may be incorrect in an unlikely scenario.
      For example, given ``It does \underline{not} \emph{release} carbon dioxide,''
      an annotator assigned a confidence score of 2 to his answer to the question \emph{What does something release?} \emph{Fresh air}.
%      in several scenarios, but there are many possibly answers and mine only accountsThe annotated answer has a chance to be true, and there might be many possible answers. For example, in ``It does \underline{not} \emph{release} carbon dioxide", the answer ``fresh air" is a possible response to the question ``What does something release?".

    \item 1: Slightly confident.
      My answer is probably correct, but there is no strong evidence in the sentence.
      For example, given ``The second plot can \underline{not} be \emph{explained} using data,''
      an annotator answered \emph{How is something explained?} with \emph{Using observations} and assigned a confidence score of~1.
      These answers often encode commonsense rather than an inference from the statement containing the target verb.

\end{compactitem}

%\tbd{\textcolor{red}{Removed a line to save some space.}}
%As we shall see (Section \ref{s:corpus-analysis}), annotators assign the highest confidence score (4/4) to most questions and
%either extremely or very confident scores (3-4/4) to almost all them.

%\begin{comment}
\begin{figure}
\centering
\includegraphics[width=0.90\columnwidth]{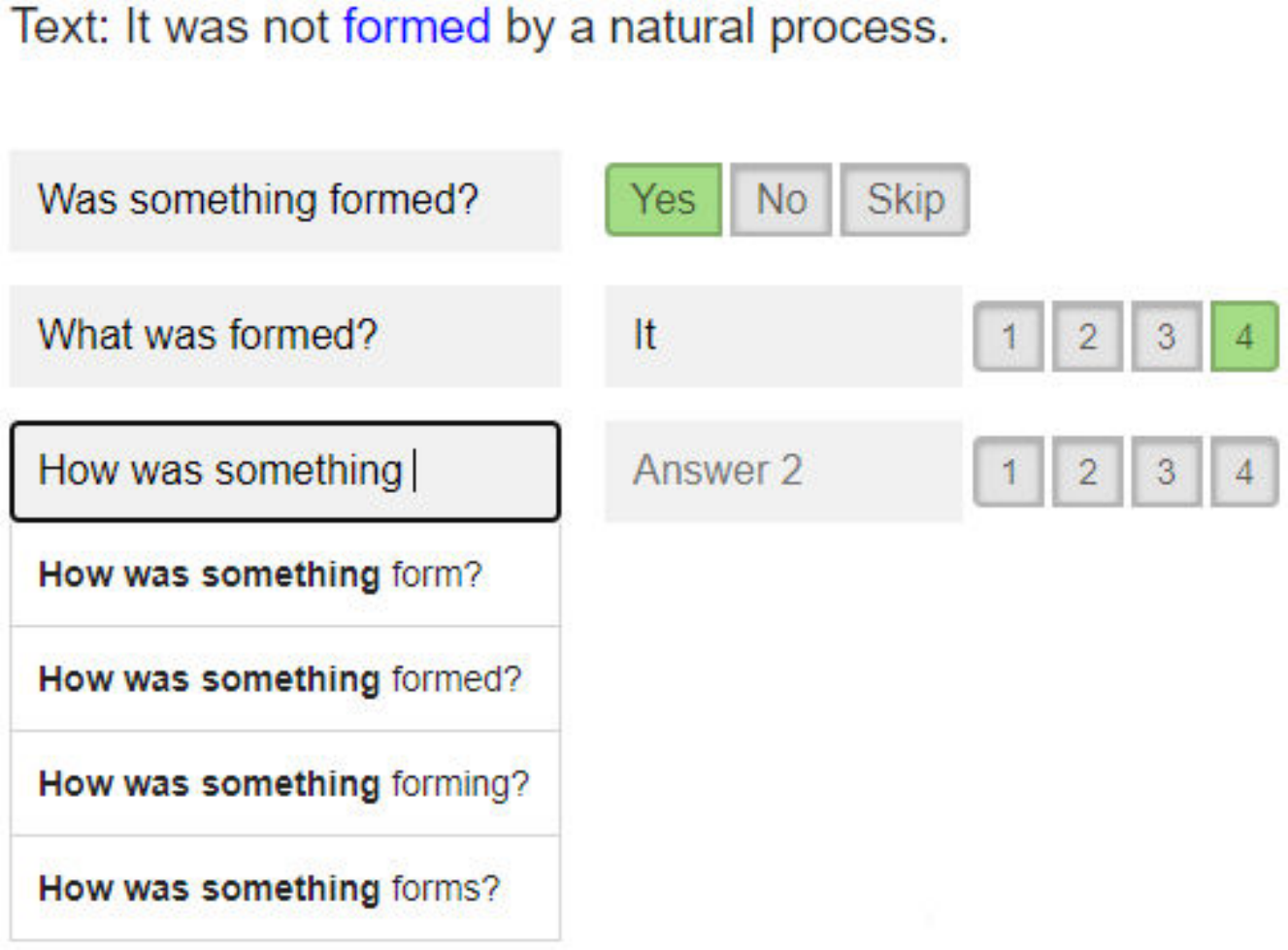} 
\caption{Web interface to guide annotators in asking and answering questions.
%  The interface seamlessly walks annotators through the template slots and makes suggestions.
  The screenshot shows the fillers for the VERB slot (i.e., the conjugation of \emph{form})
  %The numbers on the rigth indicate confidence scores.
  }
%A web interface to write questions and answers. 
%Auto-suggest shows the options for the \textbf{VRB} slot. The radio buttons shown after the answer text boxes indicate the \emph{correctness} scores. }
\label{f:auto-suggest}
\end{figure}
%\end{comment}

%Furthermore, we develop a web interface that facilitates the task of creating questions following our templates as well as answering and scoring them. Appendix \ref{s:appendix-scaling-annotation} provides details about the interface.

\noindent
\textbf{Scaling the Annotation Process}
\label{ss:auto-suggestions}
Inspired by \newcite{fitzgerald-etal-2018-large}, we develop a web interface that facilitates the task of generating questions following our templates.
More specifically, the interface auto-suggest to annotators the valid fillers for each slot.
For example, if annotators start typing \emph{W}, only fillers for the WH slot starting with \emph{W} are suggested.
The fillers for the next slot are suggested after the selection for the current slot is finalized.
Figure~\ref{f:auto-suggest} presents a screenshot of the interface with the auto-suggestions for the VERB slot (i.e., the conjugation of the target verb, \emph{form}).

\noindent
\textbf{Annotation Quality}
\label{ss:annotation-quality}
Five undergraduate students who are native English speakers participated in the annotation process.
They were trained in multiple sessions and conducted pilot annotations followed by discussion sessions before starting the annotations that resulted in the corpus described here.
We do not calculate inter-annotator agreement since two different answers to the same questions are likely to be correct.
Consider the following sentence: ``Scientists \underline{never} \emph{use} only one piece of evidence to form a conclusion.''
Two valid (and yet non-overlapping) answers to the question \emph{What does someone use?}
are 
\emph{a reasonable amount of evidence}
and
\emph{mathematical models}.
% where both ``a reasonable amount of evidence" and ``mathematical models" are valid to the question ``What does someone use?".  
The limitations of current automated metrics to determine whether these two answers are correct are well known \cite{liu-etal-2016-evaluate}, so we decided to conduct a manual evaluation.
More specifically, we manually validated 479 questions and answers 
from a random sample of 200 target verbs in the corpus.
%A sixth person not involved in the generations and answering of the questions validated the 500 question-answer pairs and discovered that only 1\% of them are incorrect.
%\tbd{\textcolor{red}{MH: I put the coefficients of both correlations since reviewers might ask for both. Showing only Spearman might be alright since scores are ordinal data type (?).}}
A sixth person not involved in the generation and answering of the questions validated the 479 question-answer pairs as well as graded them with the same 4-point confidence scale. 
The validation phase revealed that (a) only 3\% of the question-answer pairs are incorrect and (b) there is a strong correlation (Spearman: 0.71, Pearson: 0.70, p-value $<$ 0.005 for both) between the scores.
% Below are the two references for discussion of correlation strength:
% https://www.ncbi.nlm.nih.gov/pmc/articles/PMC6107969/
% https://www.statstutor.ac.uk/resources/uploaded/spearmans.pdf

\subsection{Generating Affirmative Interpretations from Questions and Answers}
\label{ss:qa-to-pi}

We devise a rule-based approach in order to go from the questions generated and answered by annotators 
to an affirmative interpretation in natural language.
Recall that annotators generate (and answer) questions in the following order:
\textit{who} (or \textit{what}) does/did  
(something) to 
\textit{whom} (or \textit{what}), 
\textit{when}, 
\textit{where}, 
\textit{how}, \textit{how much}, \textit{how many}, \textit{how long}, \textit{how often}, and
%\footnote{Along with \textbf{how}, we include a few other common options such as \textbf{how much}, how many}, \textbf{how long}, and \textbf{how often}.}, and 
\textit{why}.
Our approach is deterministic and manipulates answers depending on verb tense and number~(which are obtained with part-of-speech tags and regular expressions).

We start with the answer to the first question (who (or what) does/did something?) in order to establish the subject of the affirmative counterpart of the target verb.
Depending on whether the question uses the AUX slot, the affirmative interpretation also uses an auxiliary.
Consider the examples in Table \ref{t:qas-to-ai}.
In the first example, the question about the subject is \emph{What flows?} and the answer is \emph{Lava},
resulting in the initial affirmative interpretation \emph{Lava flows}.
Similarly, in the second example, the question is \emph{What was classified?} and the answer is \emph{Fungi},
resulting in the initial affirmative interpretation \emph{Fungi were classified}.

Having generated an initial affirmative interpretation,
the process continues adding arguments to the predicate-argument structure.
We add them sequentially to the end of the affirmative interpretation in the order in which \emph{argument questions} were generated and answered.
Consider again the first example in Table \ref{t:qas-to-ai}.
The only \emph{argument question} left is \emph{Where does something flow?},
which was answered with \emph{Close to vents}.
The initial affirmative interpretation becomes \emph{Lava flows close to vents}.
Since there are no additional questions, this is the final affirmative interpretation.
Let us now consider the second example again.
After incorporating the answer to the second question into the initial affirmative interpretation,
we have \emph{Fungi were classified as plants} (after including the preposition used in the question).
Incorporating the answer to the third question, we have the final affirmative interpretation:
\emph{Fungi were classified as plants in the past}.
%\tbd{\textcolor{red}{MH: Moved some additional details to the Appendix to save space.}}
The Appendix \ref{s:appendix-question-template} provides additional details and special cases.

\begin{comment}
The process to generate affirmative interpretations from questions and answers is robust but not foolproof from a grammatical standpoint.
Note that the semantics of the affirmative interpretation is dictated by the questions and answers,
and our evaluation determined that only \textcolor{red}{3\%} are incorrect (Section \ref{ss:annotation-quality}).
We manually validated the final affirmative interpretations for grammaticality and found that 9\% have errors.
For example, consider \emph{Most plastics do not form crystals.}
The questions and answers are as follows:
\emph{What forms something? Plastic},
\emph{What does something form? Crystals},
and
\emph{How many form? Few}.\footnote{An alternative could be to answer \emph{What forms something?} with \emph{Few plastics} (and skip the question starting with \emph{How many}).
  %This is a real example from the corpus.
  }
These result in the affirmative interpretation \emph{Plastic forms crystals few}, which places \emph{few} incorrectly.
We manually fix all the grammatical issues we found in the affirmative interpretations.
\end{comment}

\section{Corpus Analysis}
%Analyzing the Questions, Answers and Resulting Affirmative Interpretations}
\label{s:corpus-analysis}

\begin{comment}
\begin{table}[t!]
\small
\centering
\input{tables/wh-word-buckets}
\caption{Percentages of argument questions that start with each \emph{wh-word} and 
percentages of negated verbs having each \emph{wh-word}.}
\label{t:wh-word-buckets}
\end{table}
\end{comment}

\begin{table}[t!]
\small
\centering
\begin{tabular}{l cccc}
\toprule
min. confidence & $4$       &  $\ge 3$ & $\ge 2$ & $\ge 1$ \\
\midrule
\%verbs & 85.50   &  97.77       & 99.87      &  100.0  \\
\bottomrule
\end{tabular}
\caption{
Percentage of target verbs depending on the minimum confidence score assigned to any of the 
%questions generated and
answers regarding the affirmative counterpart.
Annotators almost always (97.77\%) are \emph{extremely} (4/4) or \emph{very confident} (3/4) about their answers.
}
\label{t:score_distribution}
\end{table}

\begin{table*}[t!]
\small
\centering
\begin{tabular}{l r p{5.15in}}
\toprule 
Category & \% & Example \\ 
\midrule
Patient & 24 &
  Many workers, who \emph{had} \underline{nothing} but their labour to sell, became factory workers out of necessity. $\rightarrow$ \emph{Many workers had \uwave{only their labour} to sell.} \\
Manner & 23 &
  I do\underline{n't} \emph{go} through life with regrets. $\rightarrow$ \emph{I go through life \uwave{with satisfaction}.} \\
Quantity & 10 &
  Many mutations \emph{have} \underline{no} effect on the proteins they encode. $\rightarrow$ \emph{\uwave{Some mutations} have an effect on the proteins they encode.}	 \\
Time & 10 &
  The use of asbestos is \underline{not} \emph{allowed} today. $\rightarrow$ \emph{The use of asbestos was allowed \uwave{in the past}.} \\
Reason & 9 &
  Without water, life might not be able to exist on Earth and it certainly would \underline{not} \emph{have} the tremendous complexity and diversity that we see. $\rightarrow$ \emph{Earth has complexity and diversity \uwave{because of water}.} \\
Agent & 8 &
  Unlike a resistor, an ideal capacitor does \underline{not} \emph{dissipate} energy. $\rightarrow$ \emph{\uwave{A resistor} dissipates energy.} \\
Others & 16 &
  The steep sides form because the lava can \underline{not} \emph{flow} too far from the vent. $\rightarrow$ \emph{Lava flows \uwave{close to vents}.} \\
\bottomrule
\end{tabular}
\caption{
Analysis of the arguments that differ in the target verb and the corresponding affirmative counterpart.
Categories refer to the function in the verb-argument structure.
A wavy underline indicates the new argument in the affirmative counterpart.
%The categories of verbal arguments that required changing to form affirmative interpretations. 
%\% indicates the percentage of verbs where a particular argument (e.g. Quantity) is required to change.
}
\label{t:ai-categories}
\end{table*}

The question-answer driven approach to generate and answer questions revealed that 
3,001 out of the 4,472~(negated) target verbs carry an affirmative interpretation~(67.1\%).
On average, annotators generated and answered 2.4 questions per target verb. 
Also, the average lengths of those questions and answers (in tokens) are 5.0 and 3.5, respectively. The average negated sentence is 25.8 tokens long, while its affirmative interpretation is 11.2 tokens long, indicating that affirmative interpretations are much shorter than negated sentences. 
Appendix \ref{s:additional-details-on-corpus} provides additional details, including the distribution of wh-words in the questions.
%percentage presence of each \emph{wh-word} in the corpus}.

Percentages, shown in Table \ref{t:score_distribution}, indicate that a vast majority of affirmative interpretations (85.5\%) are generated from questions and answers about which annotators were \emph{extremely confident} (confidence score: 4).
%The vast majority of affirmative interpretations are generated from answers to which annotators assigned high confidence.
%Indeed, 85.5\% of the affirmative counterparts are generated from questions and answers about which annotators were \emph{extremely confident} (Table \ref{t:score_distribution}).
The percentage raises to 97.77\% if we include questions answered about which annotators were \emph{very confident} (confidence score: 3).

%In order to analyze which arguments differ in the verb-argument structure of the (negated) target verb and the affirmative counterpart,
%we manually analyze 100 random samples from our corpus.
Similar to \newcite{sarabi-etal-2019-corpus}, we manually analyze 100 random examples from our corpus to find which arguments differ in the verb-argument structure of the (negated) target verb and the affirmative counterpart.
We discovered these arguments primarily have the following functions (Frequencies and examples in Table \ref{t:ai-categories}):

\begin{compactitem}

\item \emph{Patient (or theme)} (24\%).
The most common argument is the person or thing that is affected or acted upon by the target verb.
In the first example, we go from \emph{workers had nothing} to \emph{workers had only their labor}.
%The affirmative interpretation is formed by replacing an argument indicating a person or a thing that is affected or acted upon by the event. An example of this category is the negated sentence (4) and it's affirmative interpretation ``Many workers had only their labour to sell", where \emph{their labour} is affected by the event \emph{had}.
%Surprisingly, the highest number of the affirmative interpretations (24\%) are formed by changing the \emph{Patient} argument. 

\item \emph{Manner} (23\%).
The second most common argument is the way in which the target verb takes place (the \emph{how}).
In the example, we go from \emph{don't go through life with regrets} to \emph{go through life with satisfaction}.
%An argument expressing a manner or process needs to be changed. For example, \emph{with regrets} in sentence (5), can be changed to \emph{with satisfaction} in order to form the affirmative interpretation ``I go through life with satisfaction".

\item \emph{Quantity} (10\%).
Arguments expressing 
\emph{specific} (e.g., four, three)
or
\emph{abstract} quantities (e.g., many, less)
represent 10\% of changes in arguments.
For example, we go from 
%\tbd{\textcolor{red}{MH: replaced ``In the example" with ``For example" to save space.}}
\emph{Many mutations have no effect on the proteins}
to
\emph{Some mutations have an effect on the proteins}.
%) in the negation is replaced in the affirmative interpretation of the negation. Sentence (1) in Table \ref{t:ai-categories} is an example of \emph{abstract quantity}, where \emph{many} can be replaced with \emph{some} (or, \emph{few}) in order to generate the affirmative interpretation. 10\% of the affirmative interpretations are generated by replacing the quantities. 
%In addition, a \emph{specific quantity} can be also changed, for example, we can form an affirmative interpretation ``30 percent voters changed opinion about Obama" from the negation ``Of the respondents, 13\% viewed President Obama more favorably, while 17\% viewed him less favorably and 69\% indicated they did \underline{not} \emph{change} their opinion." In 10\% of all the events, an argument expressing quantity requires changing to construct their affirmative interpretations.

%\tbd{\textcolor{red}{MH: fixed a typo.}}
\item \emph{Time}  (10\%).
Tied in frequency with quantity, we observed arguments expressing temporal information. 
%An argument expressing a time (either \emph{specific} or \emph{abstract}) in the negated statement changes in the affirmative interpretation.
In the example, we go from
\emph{not allowed today}
to
\emph{allowed in the past}.
%An example of an \emph{abstract time} is shown in Table \ref{t:ai-categories} (Sentence 2), where the negation cue \emph{not} focuses on the argument \emph{today} which can be changed to \emph{in the past} (or, a similar phrasing indicating a time before today) to form an affirmative interpretation. 
%An example of argument expressing a specific time is ``it's premiere did \underline{not} \emph{take} place until 1988" and it's affirmative interpretation ``Its premiere took place in 1988".

\item \emph{Reason (or cause)} (9\%).
The fifth most common argument expresses the \emph{why} of the target verb.
We understand \emph{why} widely, including reasons, causes, justifications, and explanations.
In the example, we go from
something not existing \emph{without water}
to
\emph{Earth has complexity and diversity because of water}.
%An argument expressing a reason or cause in the negated sentence needs to be changed. An example of this category is the negated sentence (6) and the affirmative interpretation ``Earth has complexity and diversity because of water", where the argument \emph{without water} in sentence can be changed to \emph{because of water} to form the affirmative interpretation.

\item \emph{Agent} (8\%).
The sixth most common argument is the person or thing who performs an event (i.e., the doer).
In the example, we go from \emph{an ideal capacitor not dissipating energy} to \emph{a resistor dissipating energy}.
%The affirmative interpretation is formed by replacing an argument indicating a person or a thing who performs the event. For example, in sentence (3), the agent \emph{an ideal capacitor} needs to be replaced with \emph{a register} to form the affirmative interpretation. 

\item \emph{Other} (16\%).
Other functions (locations, purposes, recipients, etc.) account for 16\% of arguments.
Table \ref{t:ai-categories} exemplifies a location change: from
\emph{cannot flow too far from the vent}
to
\emph{flows close to the vents}.
%An argument expressing location, direction, or any other type not listed above needs changing to form an affirmative interpretation. Consider the example (7), where the argument \emph{too far from the vent} expresses a direction, and it can be changed to something such as \emph{close to vents} to form the affirmative interpretation. 
 
\end{compactitem}

\section{Experiments and Discussion}
\label{s:experiments-and-discussion}
AFIN consists of sentences containing verbal negations and their affirmative interpretations in natural language.
We experiment casting the problem of obtaining affirmative interpretations from negation
as a natural language inference task (Section \ref{s:pi-and-nli})
and as a generation task~(Section \ref{s:pi-generation}).

\subsection{Affirmative Interpretations and Natural Language Inference Classification}
\label{s:pi-and-nli}

The sentences containing the (negated) target verb and the corresponding affirmative interpretations
can be understood as the premises and hypotheses in a natural language inference (NLI) setting \cite{bowman-etal-2015-large}.
Very briefly, NLI is a classification task that determines whether a \emph{premise}
entails, is neutral with respect to, or contradicts a \emph{hypothesis}.
%As the affirmative interpretations are generated from negations, they pose an inference relationship with them. In this study, we explore 
%(1) what are the inference relations between negations and their affirmative interpretations, 
%(2) whether the state-of-the-art systems trained with existing NLI benchmarks identify the inference relations in the new benchmark, and 
%(3) whether the state-of-the-art systems fine-tuned with part of the new benchmark identify inference relations in the new benchmark? 
We label the premise-hypothesis pairs from AFIN as follows.
If all the answers to questions used to generate an affirmative interpretation received the highest confidence score (4, Extremely confident), we label them \emph{entailment} (85.5\% of the target verbs).
Otherwise (at least one answer received a confidence score between 1 and 3), we label them \emph{neutral}.
Note that contradiction examples cannot be derived from AFIN.
Here we present two examples:
%\begin{inparaenum}[(a\upshape)]
\begin{compactitem}
\item Premise: \emph{A dormant volcano no longer shows signs of activity.}
   Hypothesis: \emph{A dormant volcano showed signs of activity in the past.}
   Premise \emph{entails} hypothesis.
\item Premise: \emph{Respiratory infections such as pneumonia do not appear to increase the risk of COPD, at least in adults.}
   Hypothesis: \emph{Respiratory infections appear to increase the risk of COPD in elderly.}
   Premise is \emph{neutral} with respect to the hypothesis.
\end{compactitem}
%\end{inparaenum}

\begin{table}
\small
\centering

\begin{tabular}{cp{0.60in}  ccc c ccc}
\toprule
&                & \multicolumn{3}{c}{RoBERTa} && \multicolumn{3}{c}{XLNet} \\ 
\cmidrule{3-5} \cmidrule{7-9}
                  & Tested w/        &    P  &  R  &  F1     &&  P   &  R  &  F1 \\ 
\midrule

\multirow{3}{*}{\begin{sideways}MNLI\end{sideways}} 
                  & MNLI-dev         &  88   & 88  &  88     &&  87  & 87  & 87 \\
                  & MNLI-dev*        &  92   & 87  &  89     &&  91  & 85  & 88 \\
                  & AFIN             &  55   & 43  &  48     &&  54  & 42  & 47 \\
						
\midrule
\multirow{3}{*}{\begin{sideways}SNLI\end{sideways}} 
                  & SNLI-dev         &  92   & 92  &  92     &&  91  & 91  & 91 \\
                  & SNLI-dev*        &  93   & 90  &  92     &&  93  & 90  & 92 \\
                  & AFIN             &  56   & 37  &  45     &&  57  & 38  & 46 \\
\midrule

\multirow{2}{*}{\begin{sideways}RTE\end{sideways}} 
                  & RTE-dev          &  76   & 76  &  76     &&  70  & 68  & 69 \\
                  & AFIN             &  52   & 53  &  52     &&  53  & 55  & 54 \\

\bottomrule                        

\end{tabular}

\caption{Precision, Recall, and F1 scores (macro average) obtained with RoBERTa and XLNet
trained with MNLI, SNLI, and RTE. 
We provide results with the original development set in each benchmark,
the subsets that only contain \emph{entailment} and \emph{neutral} pairs (*),
and the premise-hypothesis pairs derived from AFIN, our corpus.
Transformers trained with any of the benchmarks perform substantially worse with AFIN.
}
\label{t:nli-results-training}
\end{table}

%\noindent
\paragraph{Transformers and Existing NLI Benchmarks}
\label{ss:training-with-existing}
%In our first experiments
At first, we seek to investigate whether state-of-the-art transformers trained with existing NLI benchmark can solve the premise-hypothesis pairs derived from AFIN.
Note that to do so, they would need to make inference in the presence of negation.
We experiment with
(a) two transformers: RoBERTa~\cite{liu2019roberta} and XLNet~\cite{yang2019xlnet},
and
(b)  three NLI benchmarks: 
MNLI~\cite{williams-etal-2018-broad},
SNLI~\cite{bowman-etal-2015-large}, and
RTE (part of the GLUE benchmark~\cite{wang-etal-2018-glue}).
We fine-tuned the transformers with the training split of each benchmark and conduct three evaluations:
with
(a)~the development split of each benchmark,
(b)~the subsets of (a) that only contain \emph{entailment} and \emph{neutral} pairs,
and
(c)~all premise-hypothesis pairs derived from AFIN.
Note that neither RTE nor AFIN have pairs annotated \emph{contradiction}. 
%Details about model training and hyperparameters are shared in the supplementary materials.
%Details about training and hyperparameters are shared in the supplementary.
%\tbd{\textcolor{red}{MH: rephrased a line}}
%Appendix \ref{ss:appendix-afin-nli} details on the training procedure and hyperparameters for the experiments.
Appendix \ref{ss:appendix-afin-nli} details the training procedure.

Table \ref{t:nli-results-training} presents the results.
While both transformers obtain roughly the same results when evaluated with the three labels or only \emph{entailment} and \emph{neutral} pairs, we observe substantial drops in F1 score when evaluated with AFIN, around 46\% with MNLI and 51\% with SNLI.

We observe a similar pattern with RTE, although the drop is relatively small with XLNet (note, however, that XLNet does much worse than RoBERTa (69 vs. 76), whose performance drops 32\%).
We hypothesize that RTE obtains better results because it does not contain \emph{contradiction} pairs.
%\tbd{\textcolor{red}{Removed the part of the sentence with colon.}}
These results show that current benchmarks are not enough to identify inferences between a negation and its affirmative interpretation.
%: transformers trained with any of them obtain results well below the majority baseline.

%\begin{comment}
\begin{table}
\small
\centering

\begin{tabular}{p{0.9in}  ccc c ccc}
\toprule
                      & \multicolumn{3}{c}{RoBERTa} && \multicolumn{3}{c}{XLNet} \\ 
\cmidrule{2-4} \cmidrule{6-8}
                      &    P  &  R  &  F1     &&  P   &  R  &  F1 \\ 
\midrule

MNLI-training         &  55   & 43  &  48     &&  54  & 40  & 46 \\
~~~+ 70\% of AFIN     &  72   & 51  &  60     &&  61  & 51  & 55 \\				
\midrule

SNLI-training         &  58   & 36  &  45     &&  60  & 38  & 47 \\
~~~+ 70\% of AFIN     &  42   & 50  &  46     &&  61  & 52  & 56 \\
\midrule

RTE-training          &  51   & 52  &  52     &&  52  & 54  & 53 \\
~~~+ 70\% of AFIN     &  56   & 53  &  54     &&  61  & 55  & 58 \\
\bottomrule                        

\end{tabular}

\begin{comment}
\begin{tabular}{l c c c}
\toprule
                      &   Maj. B.  & RoBERTa   &     XLNet   \\
\midrule
MNLI-training         &  84.55 & 53.30     &     52.10   \\
~~~+ 70\% of AFIN    &  84.55 & 84.56     &     83.89   \\
\midrule
SNLI-training         &  84.55 & 53.30     &     55.80   \\
~~~+ 70\% of AFIN    &  84.55 & 84.22     &     82.56   \\
\midrule
RTE-training          &  84.55 & 65.00     &     62.90   \\
~~~+ 70\% of AFIN    &  84.55 & 82.00     &     82.89   \\   
\bottomrule 
\end{tabular}
\end{comment}
\caption{Results obtained training with~(a)~MNLI, SNLI, or RTE
and
(b) 70\% of AFIN,
and evaluating with 30\% of AFIN.
Fine-tuning improves results, but transformers substantially underperform the original development splits (see Table \ref{t:nli-results-training}).
}
\label{t:nli-results-finetuning}
\end{table}

%\noindent
\paragraph{Fine-tuning with AFIN}
\label{ss:fine-tune-with-aineg}
The next experiments examine whether fine-tuning helps transformers identify inference relations in the premise-hypothesis pairs generated from AFIN.
To do so, we fine-tune the transformers not only with an existing benchmark (MNLI, SNLI, or RTE), but also 70\% of the pairs derived from AFIN.
Then, we evaluate with 30\% of the pairs derived from AFIN.

Table \ref{t:nli-results-finetuning} presents the results.
%train the systems with the original training split as well as with 70\% pairs from A\textsubscript{f}INeg, and then evaluate on 30\% of A\textsubscript{f}INeg. 
Perhaps unsurprisingly, fine-tuning with AFIN allows the transformers to correctly identify few more \emph{entailment} and \emph{neutral} pairs (F1 scores:~45--53 vs. 46--60).
We note, however, that no matter how we combine transformers and NLI benchmarks, the results are substantially below those obtained with the original development split (F1 scores:~46--60 vs. 69--92).
%\end{comment}

\subsection{Generating Affirmative Interpretations} % with the T5 transformer}
\label{s:pi-generation}
Casting the problem as a natural language inference task is worthwhile but unrealistic:
the affirmative interpretations to be verified (are they entailed by the sentence with the negation?)
are not readily available.
In our next experiments, we investigate a realistic formulation of the problem: \emph{generate} affirmative interpretations given a sentence with a negation.
In order to do so, we split AFIN as follows: 70\% for training, 15\% for development, and the remaining 15\% for test.

%In order to assess whether neural models can generate affirmative interpretations, we perform another experiment in which 70\% of the target events from A\textsubscript{f}INeg serve as training, 15\% as development, and the remaining as test dataset. 
%In the following discussion, we provide details about the experimental setup, results, and analysis.

\begin{table}
\small
\centering
\begin{tabular}{l ccc}
\toprule
                 & BLEU-2         & chrf++    & METEOR \\
\midrule
Negated sent.    & 26.5         & 50.5      &  43.5  \\
~~~+ target verb & 33.6         & 57.3      &  51.9  \\
\bottomrule
\end{tabular}
\caption{Evaluation results obtained with BLEU-2, chrf++, and METEOR between human and system generated affirmative interpretations.
}
\label{t:automatic-metrics}
\end{table}

\noindent
\textbf{Experimental Setup}
\label{s:experimental-setup}
We perform the experiments with the T5-Large transformer~\cite{2020t5}, which can generate text through a supervised learning setup. 
In particular, we train T5 to generate affirmative interpretations using two inputs:
(a) only the sentence containing the (negated) target verb (i.e., the negated sentence), and %\tbd{\textcolor{red}{Rewrote a sentence here.}}
(b) the negated sentence concatenated with the target verb. 
%In the first setup, we investigate whether T5 can identify the negation and generate an affirmative interpretation.
%\tbd{\textcolor{red}{Removed a sentence and then rewrote the next sentence.}}
The second setup investigates whether inputting the target verb with the negated sentence aids in generating affirmative interpretation about that target verb.
%We provide details about the training procedure and hyperparameters in the supplementary materials.
%\tbd{\textcolor{red}{MH: rephrased a line}}
Additional details on the training procedure are provided in Appendix \ref{ss:appendix-afin-generation}.

\noindent
\textbf{Results and Analysis}
\label{s:results-and-analysis}
After the training process with both setups, 
we obtain evaluation scores using three automatic metrics:
BLEU-2 \cite{papineni2002bleu}, 
chrf++ \cite{popovic2017chrf++}, and 
METEOR \cite{banerjee2005meteor}.
We calculate these metrics comparing the human- and T5-generated affirmative interpretations from the test split (Table~\ref{t:automatic-metrics}).
%between human and T5 generated affirmative interpretations for the negations from the test split (Table~\ref{t:automatic-metrics}).
Evidently, the system provided with the target verb shows comparatively better scores than the system without the target verb (e.g., BLEU-2: 33.6 vs 26.5). 
%\tbd{\textcolor{red}{Interpreting the scores seem little challenging to me. }} 
Based on the scores from the best setting, T5 achieves some capability to automatically generate affirmative interpretations.

While useful, automatic metrics
only provide a partial picture about the quality of affirmative interpretations, as outlined in Section \ref{ss:annotation-quality}.
%might only serve as weak evaluation strategies due to the reasons outlined in Section \ref{ss:annotation-quality}. 
Therefore, we manually evaluate the affirmative interpretations generated by T5. 
%Manual evaluation is sound because of the reasons described in Section \ref{ss:annotation-quality}.
In particular, the same annotator that validated a sample of AFIN validated the output of T5 with the confidence scores provided in Section \ref{ss:annotation-quality}.\footnote{The only difference is that the affirmative interpretations come from T5 instead of a human annotator.}
Note that this time we added a new score of 0 to indicate that an affirmative interpretation is incorrect.
%We added a new score of 0 to indicate that the affirmative interpretation is incorrect.
%Note that this score does not apply when annotators assign a confidence score to their answers.

%T5 generates affirmative interpretations for the instances on the test split for the above two setups. 
%Because of the reasons explained in Section \ref{ss:annotation-quality}, we evaluate the generated affirmative interpretations manually instead of using any automatic metric. 
%In particular, a student annotator grades the affirmative interpretations generated by T5, with the same confidence labels discussed in Section \ref{sss:correctness-scoring}, along with a score 0 (as \emph{not true}) to account the incorrect cases.
%Table \ref{t:generated-ai} compares the quality of affirmative interpretations generated by humans and those generated by T5.
%In case of first row (\emph{Humans}), if a question is answered with a lower score (e.g., 2) of an event, then the generated affirmative interpretation is assigned to that score (e.g., 2). 

Table \ref{t:generated-ai} provides the results.
The scores assigned to AFIN represent an upper bound.
We observe that explicitly providing the (negated) target verb is beneficial as it allows T5 to generate many more \emph{extremely confident} affirmative interpretations (32\% vs. 43.3\%).
We observe, however, that T5 faces challenges generating affirmative interpretations.
First, over a quarter (27.3\%) are incorrect.
Second, compared to AFIN (i.e., human annotators),
T5 only generates about half (43.3\% vs. 86.2\%) of affirmative interpretations that an evaluator is \emph{extremely confident} about (confidence score: 4).

\begin{table}
\small
\centering
\begin{tabular}{l r r r r r}
\toprule
     & \multicolumn{5}{c}{Confidence Scores} \\ \cmidrule{2-6}
     &    \multicolumn{1}{c}{4}    &   \multicolumn{1}{c}{3}     &   \multicolumn{1}{c}{2}     &   \multicolumn{1}{c}{1}   &   \multicolumn{1}{c}{0}   \\ 
\midrule
AFIN (upper bound)        &   86.2  &   11.6  &   2.0    &   0.2  &  n/a    \\
\addlinespace
T5-Large \\
~~~~Negated sent.         &   32.0  &   15.3  &   12.0   &   3.3   &   37.3  \\
~~~~~~~~+ target verb     &   43.3  &   10.0  &   15.3   &   4.0   &   27.3  \\
\bottomrule

\end{tabular}
\caption{Percentage of affirmative interpretations assigned each confidence score in (a) the AFIN test set
and (b) those generated by T5 (not providing and providing the target verb).
T5 substantially underperforms AFIN, which is a human upper bound.
%
%)Comparison of the quality of affirmative interpretations between what humans generate and what T5-Large generates on the test split. 
%Numbers are in percentage, indicating the \% of affirmative interpretations under each confidence score.
}
\label{t:generated-ai}
\end{table}

\begin{figure}
\centering
\includegraphics[width=0.90\columnwidth]{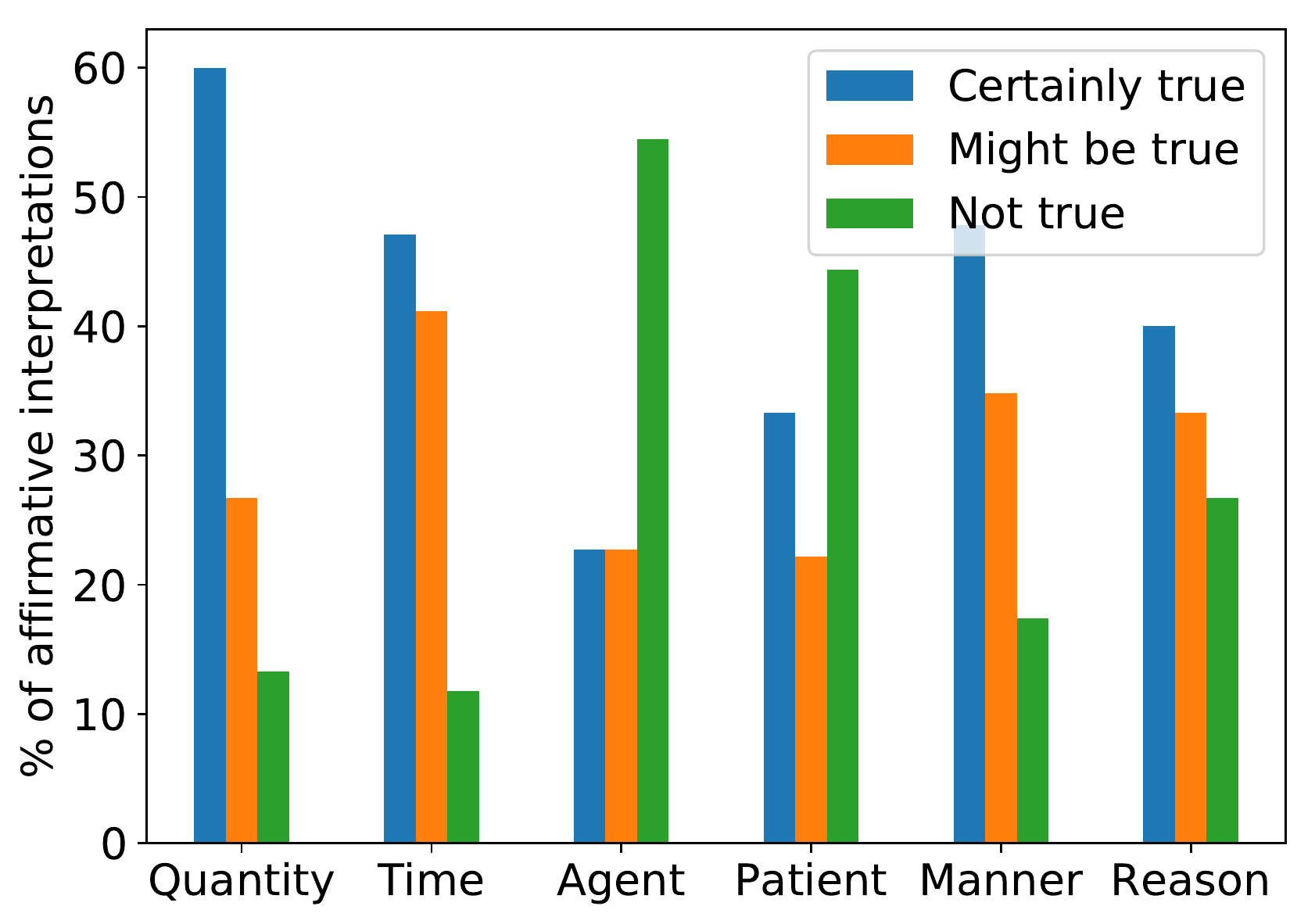} 
\caption{
%Confidence score analysis of the affirmative interpretations generated by T5.
  %(Certainly true: 4, Might be true: 1--3, Not true: 0).
Analysis of scores assigned to the affirmative interpretations generated by T5. 
Scores are much lower when the argument that has to be changed to generate the affirmative interpretation is an \emph{agent} or \emph{patient}.
  %All but \emph{quantity} have less than 50\% certainly true.
% of \emph{certainly true} (score: 4), \emph{might be true} (score: 1-3), and \emph{not true} (score: 0) affirmative interpretations under each category. 
%Note that \emph{probably true}, \emph{possibly true}, and \emph{least likely true} responses are combined into a new label (i.e., \emph{might be true}) for better visualization. 
}
\label{f:cat-error-dist}
\end{figure}

\noindent
\textbf{Qualitative Analysis}
In addition to confidence scores, we also analyze when T5 faces the biggest challenges generating affirmative interpretations.
To this end, we randomly selected 150 instances from the test split.
Then, we manually annotated
the functions of the arguments that should be replaced in the affirmative interpretations
with the same categories than the ones discussed in Section \ref{s:corpus-analysis}.
We present the confidence score analysis in Figure \ref{f:cat-error-dist}.
%To analyze when the best system struggles, we randomly select 150 negations from the test split, then manually annotate them with the same categorical labels as in Section \ref{s:corpus-analysis}. 
For convenience, we show scores in three groups: 
\emph{certainly true} (score: 4),
\emph{might be true} (scores from 1 to 3), and
%we merged \emph{probably true}, \emph{possibly true}, and \emph{least likely true} cases) and,
\emph{not true} (score: 0).
%under each argument label (Figure \ref{f:cat-error-dist}). 

We observe that it is comparatively easy for T5 to generate \emph{certainly true} affirmative interpretations when the argument to be replaced contains a quantity (\emph{certainly true} vs. \emph{not true}: 60\% vs 13.3\%). 
%\tbd{\textcolor{red}{MH: Replaced ``In other words" with ``Therefore" to save some space.}}
Therefore, T5 learned some patterns to replace quantities in the affirmative interpretations.
%For example, 
%This is because the system might struggle less to infer a replacement required for the \emph{quantity} argument in order to generate the affirmative interpretation. 
For example, from negation 
``Schools can \underline{not} \emph{charge} students more than US\$5 to defray the cost of insurance,''
T5 correctly generates 
``Schools can charge students US\$5 to cover the cost of insurance.'' 
%Which indicates the system finds less hard to infer a replacement such as ``US\$ 5" for the \emph{quantity} argument ``more than US\$ 5".
%In case of the \emph{abstract quantity} type, the system also able to infer easy opposites. 
%For example, for the negation ``Many climate types have \underline{not} been \emph{mentioned} here", the system generates ``Some climate types were mentioned here", where the quantity \emph{many} is changed to \emph{some} in the generation process.
Despite the relatively success with quantities, 
%\tbd{\textcolor{red}{MH: removed text ``(60\% affirmative interpretations are certainly true)". Same information is already available in a few lines above}}
%(60\% affirmative interpretations are certainly true), 
less than 50\% of all affirmative interpretations that require replacing an argument in any other category are deemed certainly correct.
\emph{Agent} and \emph{patient} are the categories T5 finds most challenging---these affirmative interpretations are more often deemed \emph{not true} than \emph{certainly true}.
T5 often generates affirmative interpretations in these categories by deleting the negation cue and fixing verb tense and auxiliaries to form a grammatical---but incorrect---affirmative interpretation.
For example, given ``Ryanair have also sacked veteran pilot John Goss for appearing on the show, the only pilot interviewed who did \underline{not} \emph{seek} anonymity,''
T5 generates ``Veteran pilot John Gosson sought anonymity.''

\section{Conclusions}
\label{s:conclusions}
We have proposed a question-answer driven approach to reveal affirmative interpretations from verbal negations.
Annotators generate and answer questions regarding the affirmative counterpart of a negated verb,
and then we generate from them an affirmative counterpart in natural language.
Through analyses, we have shown that 67.1\% of verbal negations convey that the negated event is actually factual.
More importantly, we observe many categories in the arguments that are replaced in the affirmative interpretations (patient, manner, quantity, time, reason, etc.).
%The experiments show that casting the problem as NLI does not outperform the majority baseline. 
The experiments show that transformers struggle substantially when we cast the problem as NLI.
Doing so, however, is an unrealistic scenario: affirmative interpretations are not readily available to be fed into a natural language inference classifier.
Further, we observe very limited success
generating affirmative interpretations given as input a sentence containing verbal negation.
We argue that generating affirmative interpretation is the realistic scenario
and propose doing so as a challenging generation task requiring a combination of language comprehension, commonsense, and world knowledge currently out of reach for state-of-the-art models.

\section*{Acknowledgements}
This material is based upon work supported by the National Science Foundation under 
Grant No.~1845757.
Any opinions, findings, and conclusions or recommendations expressed in this material
are those of the authors and do not necessarily reflect the views of the NSF.
The Titan Xp used for this research was donated by the NVIDIA Corporation.
Computational resources were also provided by the UNT office of High-Performance Computing. 
Additionally, we utilized computational resources from the Chameleon platform \cite{keahey2020lessons}. 
We also thank the anonymous reviewers for their insightful comments.

\bibliography{refs}
\bibliographystyle{acl_natbib}

\appendix
\section{Additional Details on Template-based Question Generation}
\label{s:additional-question-template}

This section provides additional details for the slots in the template-based question generation presented in Section \ref{ss:annotation-task-design} of the paper.

\begin{compactitem}
\item WH indicates wh-words to generate the argument questions. 
The complete set of options we use are as follows: \emph{who, what, whom, when, where, how, how much, how many, how long, how often}, and \emph{why}.
\item AUX indicates auxiliary verbs. 
The predicate questions always start with an auxiliary. 
However, argument questions may or may not contain an auxiliary verb (See examples in Table 3 of the paper). 
We avail the below list of auxiliary verbs for annotators: \emph{is, was, does, did, has, had, can, could, may, might, will, would, should}, and \emph{must}.
\item SUB refers to subjects. 
Similar to \newcite{he-etal-2015-question}, we only avail \emph{someone} or \emph{something}, indicating placeholder for the subject position.
\item VERB indicates the full conjugation of the target verb.
\item OBJ1 refers to the options for objects. 
Similar to SUB, we only avail \emph{someone} or \emph{something}, indicating placeholders for objects.
\item PREP refers to prepositions. We avail a short list of common prepositions: \emph{by, to, for, with, about, of}, and \emph{from}.
\item OBJ2 refers to the additional options for objects. The complete list includes the following: \emph{someone, something, somewhere, do, doing, do something}, and \emph{doing something}.
\end{compactitem}

\begin{comment}
\section{Scaling the Annotation Process}
\label{s:appendix-scaling-annotation}

As mentioned in Section \ref{sss:correctness-scoring}, we build a web interface inspired by \newcite{fitzgerald-etal-2018-large}, which facilitates the process of generating questions following our templates.
More specifically, the interface auto-suggests to annotators the valid fillers for each slot.
For example, if annotators start typing \emph{W}, only fillers for the WH slot starting with \emph{W} are suggested.
The fillers for the next slot are suggested after the selection for the current slot is finalized.
Figure~\ref{f:auto-suggest} presents a screenshot of the interface with the auto-suggestions for the VERB slot (i.e., the conjugation of the target verb, \emph{form}).
\end{comment}

\begin{comment}
\begin{figure}
\centering
\includegraphics[width=0.75\columnwidth]{figures/auto-suggest} 
\caption{Web interface to guide annotators in asking and answering questions.
%  The interface seamlessly walks annotators through the template slots and makes suggestions.
  The screenshot shows the fillers for the VERB slot (i.e., the conjugation of \emph{form}).
  %The numbers on the rigth indicate confidence scores.
  }
%A web interface to write questions and answers. 
%Auto-suggest shows the options for the \textbf{VRB} slot. The radio buttons shown after the answer text boxes indicate the \emph{correctness} scores. }
\label{f:auto-suggest}
\end{figure}
\end{comment}

\begin{table*}[th!]
\small
\centering
\begin{tabular}{l cccccc p{1.0in} p{1.5in}}
\toprule

                     & WH         &   AUX   &   SUB        &  VERB     &      PREP  & &  Answer & Affirmative Interpretation \\ \midrule
                                          
\multirow{2}{*}{(1)} & What       &         &              &  happens   &            &  ?   & Reflection  & Reflection happens.  \\ 
& What      &   does  &  something   &  happen    &     with       &  ?   & Any type of waves     & Reflection happens with any type of waves. \\ 
\midrule

\multirow{2}{*}{(2)} & What       &     was    &             &  made   &    by        &  ?   & It  & It was made.  \\ 
& What      &   was  &  something   &  made    &     by        &  ?   & Inanimate organisms     & It was made by inanimate organisms. \\ 
%&            &         &              &          &            &      &                    &  \\ 
\midrule  

\multirow{2}{*}{(3)} & What       &     has    &    something          &     &            &  ?   & Later life forms  & Later life forms have.  \\ 
& What      &   does  &  something   &  have    &            &  ?   & The ability to photosynthesize     & Later life forms have the ability to photosynthesize. \\ 
%&            &         &              &          &            &      &                    &  \\ 
\midrule  

\multirow{4}{*}{(4)} & What       &      &              &  rises   &        &  ?   & The Sun             & The Sun rises. \\ 
& When       &  does    &  something   &  rise   &      &  ?   & In all seasons            & The Sun rises in all seasons. \\ 
& Where      &  does    &  something   &  rise   &      &  ?   & In the sky       & The Sun rises in all seasons in the sky. \\
& How much      &  does    &  something   &  rise   &      &  ?   & Very low       & The Sun rises in all seasons in the sky very low. \\
\midrule  

\multirow{3}{*}{(5)} & Who       &      &              &  returned   &        &  ?   & Locke             & Locke returned. \\ 
& When       &  did    &  someone   &  return   &      &  ?   & After the Glorious Revolution            & Locke returned after the Glorious Revolution. \\ 
& Where      &  did    &  someone   &  return   &      &  ?   & Home       & Locke returned after the Glorious Revolution home. \\

\bottomrule
\end{tabular}
\caption{Examples of questions and answers generated by annotators and the resulting affirmative interpretations.
The sentences containing the negated predicates are
(1) \emph{Reflection can \emph{happen} with any type of waves, \underline{not} just sound waves},
(2) \emph{It was \underline{not} \emph{made} by living organisms},  
(3) \emph{The earliest life forms did \underline{not} \emph{have} the ability to photosynthesize}, 
(4) \emph{Even in summer, the Sun \underline{never} \emph{rises} very high in the sky}, and 
(5) \emph{Locke did \underline{not} \emph{return} home until after the Glorious Revolution}.
We do not show the OBJ1 and OBJ2 slots because they are empty for the questions in these examples.
}
\label{t:qas-to-ai-more}
\end{table*}

\section{Additional Details on Generating Affirmative Interpretations from Questions and Answers}
\label{s:appendix-question-template}

The process to generate affirmative interpretations from questions and answers is robust but not foolproof from a grammatical standpoint.
Note that the semantics of the affirmative interpretation is dictated by the questions and answers,
and our evaluation determined that only 3\% are incorrect (Section \ref{ss:annotation-quality}).
We manually validated the final affirmative interpretations for grammaticality and found that 9\% have errors.
For example, consider \emph{Most plastics do not form crystals.}
The questions and answers are as follows:
\emph{What forms something? Plastic},
\emph{What does something form? Crystals},
and
\emph{How many form? Few}.\footnote{An alternative could be answering \emph{What forms something?} with \emph{Few plastics} (and skip the question starting with \emph{How many}).
  %This is a real example from the corpus.
  }
These result in the affirmative interpretation \emph{Plastic forms crystals few}, which places \emph{few} incorrectly.
We manually fix all the grammatical issues we found in the affirmative interpretations.
Table \ref{t:qas-to-ai-more} provides additional examples (Similar to Table \ref{t:qas-to-ai} in the paper).

\section{Additional Details on Corpus Analysis}
\label{s:additional-details-on-corpus}

Table \ref{t:length-buckets} presents percentages of negated sentences and their affirmative interpretations in several length buckets.
In the corpus, sentences with negation are fairly long, for example, 29.76\% of them are longer than 29 tokens. 
The affirmative interpretations, however, are much shorter, with 79.9\% being under 15 tokens. 

In Table  \ref{t:wh-word-buckets}, we report the percentage of the argument questions that start with each wh-word (first column) and the percentage of negated verbs that contain a wh-word (second column). For example, 52.61\% of all the argument questions start with the wh-word \emph{what}, and 92.27\% of all the negated verbs contains at least one question that starts with \emph{what}.

%\begin{comment}
\begin{table}[t!]
\small
\centering
\begin{tabular}{l r r}
\toprule
Lengths         &    \%Neg. Sentences   & \%Affirm. Interpretations \\ \midrule

$<$10    &    4.10         &   43.12 \\ 
10--14   &    13.93        &   36.79 \\ 
15--19   &    21.09        &   13.53 \\ 
20--24   &    19.06        &   4.30   \\
25--29   &    12.06        &   1.53  \\
$>$29    &    29.76        &   0.73  \\ 
 \midrule
All      &    100          &   100  \\ 
\bottomrule
\end{tabular}
\caption{Percentages of negated sentences and affirmative interpretations in different length buckets.
Length is measured in tokens. 
The average length of a negated sentence and its affirmative interpretation is 25.8 and 11.2, respectively. }
\label{t:length-buckets}
\end{table}
%\end{comment}

\begin{table}[t!]
\small
\centering
\begin{tabular}{l r r}
\toprule
          &    \%          &   \%verb with \\ 
\midrule

What      &    52.61        &   92.27 \\ 
Who       &    17.27        &   39.19 \\ 
when      &    9.89         &   23.89 \\ 
how       &    7.09         &   17.09 \\
where     &    6.31         &   15.19 \\
why       &    3.60         &   8.73  \\ 
how much  &    1.65         &   4.00  \\
how often &    0.63         &   1.53  \\
how many  &    0.51         &   1.23  \\ 
how long  &    0.41         &   1.00  \\ 
whom      &    0.03         &   0.07  \\
\bottomrule
\end{tabular}
\caption{Percentages of argument questions starting with each wh-word and 
percentages of negated verbs containing questions that start with each wh-word.}
\label{t:wh-word-buckets}
\end{table}

\section{Training Procedure and Hyperparameters}
\label{s:appendix-training-hyperparameters}

\subsection{Affirmative Interpretations and Natural
Language Inference Classification}
\label{ss:appendix-afin-nli}

For all the experiments mentioned in Section \ref{s:pi-and-nli} in the paper, we use Huggingface implementation \cite{Wolf2019HuggingFacesTS} of the transformer systems. 
In addition, we utilize the base architecture (12-layer, 768-hidden, 12-heads) of transformers and their pretrained weights. 
We accept the default setting for most of the hyperparameters, except a few carefully selected to fine-tune the systems. 
Table \ref{t:nli-hyperparams} shows the hyperparameters used to fine-tune RoBERTa~\cite{liu2019roberta} and XLNet~\cite{yang2019xlnet} on the three NLI corpora. Our code is available at \url{https://github.com/mosharafhossain/AFIN}.

\begin{table*}[th!]
\small
\centering
\begin{tabular}{lc ccc ccc ccc}
\toprule
\multirow{2}{*}{Hyperparameter}  &&
 \multicolumn{2}{c}{RTE}  &&
 \multicolumn{2}{c}{SNLI} &&
 \multicolumn{2}{c}{MNLI} \\
 \cmidrule(lr){3-4} \cmidrule(lr){6-7} \cmidrule(lr){9-10}
&& RoBERTa & XLNet && RoBERTa & XLNet && RoBERTa & XLNet \\
\midrule

Batch size         && 16 & 8 
                   && 32 & 32 
                   && 32 & 32 \\
Learning rate      && 2e-5 & 2e-5 
                   && 1e-5 & 1e-5
                   && 2e-5 & 2e-5 \\
Epochs             && 10 & 50 
                   && 3 & 3
                   && 3 & 3 \\
Weight decay       && 0.0 & 0.0 
                   && 0.1 & 0.1
                   && 0.0 & 0.0 \\                   
\bottomrule
\end{tabular}

\caption{Hyperparameters for finetuning the transformer systems used in Section \ref{s:pi-and-nli} in the paper.
}
\label{t:nli-hyperparams}
\end{table*}

\subsection{Generating Affirmative Interpretations}
\label{ss:appendix-afin-generation}

In order to generate affirmative interpretations for both input configurations (Section \ref{s:pi-generation} in paper), we use the same set of hyperparameters discovered through cross-validation to tune the T5-Large system. Further, for the setup that adds the target verb with the sentence containing negation, we use two prefixes\footnote{https://huggingface.co/docs/transformers/model\_doc/t5} (one for the \emph{target verb} and another for the \emph{negated sentence}) to create a single text before encoding it and passing to the T5 system. 
During the training process, we stop as soon as the loss (T5 uses \emph{cross-entropy}) in the development split does not increase for 10 epochs. Thus, the final model is the one that produces the lowest loss in the development split.
Table \ref{t:params-generation} provides the list of hyperparameters values in our experiments. 
In each run, the model requires approximately three hours to train on a single NVIDIA Tesla K80 GPU.
The code is available at \url{https://github.com/mosharafhossain/AFIN}.

\begin{table}[th!]
\small
\centering
\begin{tabular}{lc}
\toprule
Hyperparameter           \\
\midrule
Max Epochs        &  50 \\
Batch Size        &  4 \\
Sentence max length   & 128 \\
Optimizer         & Adafactor \\
Learning rate     & 1e-5 \\
Weight decay    & 5e-6  \\
Warmup epoch    & 5 \\
Accumulate step & 1 \\
Grad\_clipping   & 5.0 \\
Top\_k           & 50 \\
Top\_p           & 0.95 \\
Repetition\_penalty  & 2.5 \\
\bottomrule

\end{tabular}

\caption{Hyperparameters for finetuning T5-Large on AFIN (Section \ref{s:pi-generation} in paper). 
}
\label{t:params-generation}
\end{table}

\end{document}